\pdfoutput=1

\documentclass[11pt]{article}

\usepackage[preprint]{acl}

\usepackage{times}
\usepackage{latexsym}

\usepackage[T1]{fontenc}

\usepackage[utf8]{inputenc}

\usepackage{microtype}

\usepackage{inconsolata}

\usepackage{graphicx}

\usepackage{hyperref}
\usepackage{color}

\usepackage{standalone}
\usepackage{tikz}
\usetikzlibrary{shapes.geometric, arrows.meta, positioning, calc}
\tikzset{
font={\fontsize{9pt}{9pt}\selectfont},
  box/.style={rectangle, rounded corners, draw, fill=#1, align=center, inner sep=4pt},
  arrow/.style={-{Latex[scale=1.0]}, thick},
}

\usepackage{todonotes}
\usepackage[most]{tcolorbox}
\usepackage{tabularx, booktabs, makecell, ragged2e}
\newtcolorbox{promptbox}[2][]{text width=\columnwidth, enhanced jigsaw, breakable, size=title,
colback=black!5!white, colframe=gray!75!black, fonttitle=\bfseries,
title=#2, pad at break=1mm, break at=12.8cm/0pt, left=2pt, right=0pt}
\newcommand{\darkred}[1]{\textcolor{red!80!black}{#1}}
\newcommand{\darkgreen}[1]{\textcolor{green!70!black}{#1}}
\newcommand{\tikzxmark}{%
\tikz[scale=0.28] {
    \draw[red, line width=1.5,line cap=round] (0,0) to [bend left=6] (1,1);
    \draw[red, line width=1.5,line cap=round] (0.2,0.95) to [bend right=3] (0.8,0.05);
}}
\newcommand{\tikzcmark}{%
\tikz[scale=0.28] {
    \draw[green!90!black, line width=1.5,line cap=round] (0.25,0) to [bend left=10] (1,1);
    \draw[green!90!black, line width=1.5,line cap=round] (0,0.5) to [bend right=1] (0.23,0);
}}

\title{Language Models Benefit from\\Preparation with Elicited Knowledge}

\author{Jiacan Yu \\
  University of Rochester\\
  \texttt{jyu48@u.rochester.edu} \\\And
  Hannah Y. An \\
  University of Rochester \\
  \texttt{yan2@cs.rochester.edu} \\\And
  Lenhart K. Schubert \\
  University of Rochester \\
  \texttt{schubert@cs.rochester.edu} \\}

\begin{document}
\maketitle
\begin{abstract}
The zero-shot chain of thought approach \citep{NEURIPS2022_8bb0d291} is often used in question answering (QA) by a language model (LM) for tasks that require multiple reasoning steps. However, some QA tasks hinge more on accessing relevant knowledge than on chaining reasoning steps. We introduce a simple prompting technique, called PREP, that involves using two instances of LMs: the first (LM1) generates relevant information, and the second (LM2) receives the information from the user and answers the question. This design is intended to make better use of the LM's instruction-following capability. PREP is applicable across various QA tasks without domain-specific prompt engineering. To develop PREP, we create a dataset of 100 QA questions derived from an extensive schematic dataset specifying artifact parts and material composition. These questions ask which of two artifacts is less likely to share materials with another artifact. Such questions probe the LM's knowledge of shared materials in the part structure of different artifacts. We test our method on our shared material dataset and three published commonsense reasoning datasets. The average accuracy of our method is consistently higher than that of all the other tested methods across all the tested datasets.

\end{abstract}

\section{Introduction}
Zero-shot Chain of Thought (CoT) \citep{NEURIPS2022_8bb0d291} has emerged as a popular technique in question answering (QA) by a language model (LM), particularly for tasks that require multiple reasoning steps. This approach leverages the prompt ``Let's think step by step.'' to enhance the model's capacity for handling complex reasoning processes. However, not all QA tasks benefit from this method.
\citet{NEURIPS2022_8bb0d291} show that the zero-shot CoT approach does not offer any performance improvements on commonsense reasoning datasets, particularly CSQA \citep{talmor2019commonsenseqaquestionansweringchallenge} and StrategyQA \citep{Geva2021DidAU}. 
This suggests that many tasks rely more on the model's ability to effectively access and utilize relevant knowledge than on step-by-step reasoning.

To address this limitation, we introduce a simple general prompting technique, PREP, that involves using two instances of an LM. In our approach, the first instance (LM1) generates information relevant to answering the question, and the second instance (LM2) uses this information to answer the question. PREP is designed to be applicable to a wide range of QA tasks without requiring specialized prompt engineering.
PREP is designed for instruction fine-tuned LMs. It involves copying the information provided by LM1 and sending it to LM2 as part of the user's instruction, rather than asking the same instance of LM to answer the question after it retrieves the information. This makes better use of the LM's instruction-following capability.



We developed our prompting method on a curated dataset of 100 binary-choice questions. These questions are derived from an extensive schematic dataset presented in \citet{an2024largelanguagemodelstool}, which includes detailed information on the parts and material composition of various artifacts. The binary-choice questions ask LMs to determine which of two artifacts is less likely to share materials with another artifact, thus challenging the models to utilize detailed partonomic and material knowledge. While this is a subdomain of commonsense knowledge, it is an important one, since much of our knowledge about everyday objects concerns their part structure and material makeup. For example, if someone were asked to describe the essential characteristics of a pencil, the answer would probably include mention of the wooden shaft, graphite core, and protruding graphite point (besides usage, of course). Some depth of knowledge is required to recognize that a pencil shares some of its material composition with lithium-ion batteries -- whose anodes are typically coated with graphite.

In our experiments, we test PREP on these questions and questions from published commonsense reasoning datasets. 
The results demonstrate the efficacy of our dual-instance prompting approach. 
To the best of our knowledge, our method is one of the simplest approaches for improving commonsense QA performance that does not require human-crafted examples.

\section{Related Work}
\subsection{Related Work on Generic User-independent LM Prompting}

The zero-shot Chain of Thought method, introduced by \citet{NEURIPS2022_8bb0d291}
involves a two-stage process. First, it elicits step-by-step reasoning from the model using a general trigger phrase: ``Let's think step by step.'' Second, it extracts the final answer from the generated reasoning. This approach allows the model to perform complex multi-step reasoning tasks by mimicking the human thought process, improving performance across various benchmarks, including arithmetic and reasoning tasks.

While zero-shot CoT focuses on enhancing the model's ability to answer questions that are expected to require step-wise inference of intermediate results (for example, to derive numerical answers), our method aims to improve performance on tasks that require extensive factual knowledge.

Similar to zero-shot CoT, Plan-and-Solve (PS) prompting, proposed by \citet{Wang2023PlanandSolvePI}, replaces the trigger phrase ``Let's think step by step'' with a longer phrase that instructs the LM to devise and execute a plan. This approach is intended to address some limitations of zero-shot CoT, such as calculation errors, missing-step errors, and semantic misunderstandings.

\subsection{Prompting Methods that Require Task-Specific Design}

Perhaps the method most similar to our own is generated-knowledge prompting \citep{liu2022generated}. This prompts for knowledge relevant to a multiple-choice question, after providing five hand-engineered examples of question-knowledge pairs for each of the four commonsense domains from which the question is taken: NumerSense (focused on numeric attributes, such as numbers of certain parts of an entity, or the freezing temperature of a certain liquid); CSQA (crowd-sourced questions and answers inspired by ConceptNet pairs of triples such as \textit{stream AtLocation river, river AtLocation bridge}); CSQA2 (also based on ConceptNet-inspired questions, obtained via a QA game); and QASC (grade school science questions, e.g., about using wind to generate electricity). Our method differs in that it does not require engineering prompts tailored to specific classes of questions. In a somewhat similar vein, \citet{wang2022iterative} developed a Transformer-based iterative prompter designed for multi-hop reasoning, which was trained on crowd-sourced multi-hop datasets. Again, our approach is less demanding, as it eliminates the need for training or downloading a separate prompting LM.

Some other works in the field make use of more sophisticated reasoning processes, such as tree of thoughts (ToT) \citep{10.5555/3666122.3666639} and graph of thoughts (GoT) \citep{besta2024graph}, to improve performance. These methods require significant human intervention to design task-specific mechanisms that shape the reasoning process into a tree or graph structure.
The ToT framework enhances the problem-solving abilities of LMs by structuring the reasoning process as a tree. Each node in this tree represents a partial solution. To apply this method to a task, a user needs to determine a general strategy for decomposing intermediate thought steps based on the task properties. The method also requires a thought generator to propose potential next partial solutions, and a state evaluator to assess their effectiveness. Then search algorithms like breadth-first search (BFS) and depth-first search (DFS) are used to explore and refine potential solutions. ToT has demonstrated improvements in solving complex tasks such as the Game of 24, creative writing, and mini crosswords.


\section{Our PREP Approach}

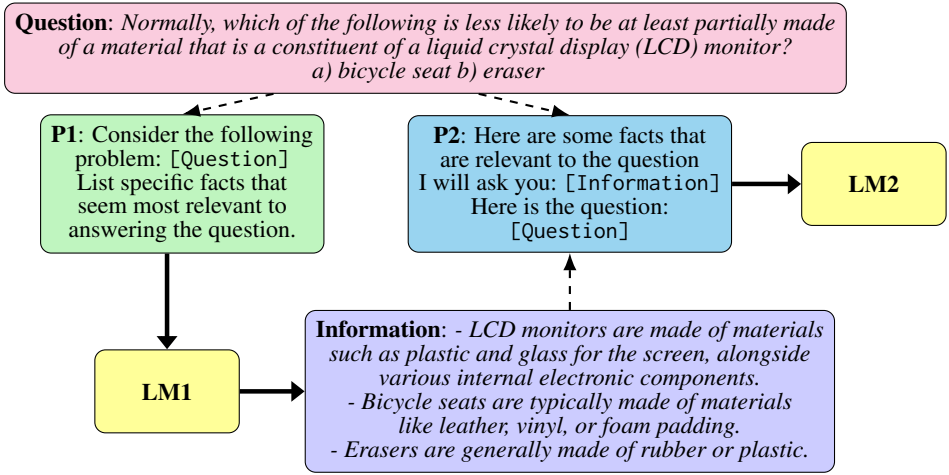
\begin{figure*}[t!]
    \centering
    \begin{tikzpicture}[node distance=0.5cm and 1cm]

\definecolor{green}{rgb}{0.75, 0.96, 0.75}
\definecolor{pastelblue}{rgb}{0.6, 0.83, 0.94}
\definecolor{pink}{rgb}{0.98, 0.8, 0.87}
\definecolor{purple}{rgb}{0.8, 0.8, 1}
\definecolor{pastelyellow}{rgb}{1, 1, 0.6}

\node (q) [box=pink, minimum height=0.6cm] {\textbf{Question}: \textit{Normally, which of the following is less likely to be at least partially made}\\
\textit{of a material that is a constituent of a liquid crystal display (LCD) monitor?}\\
\textit{ a) bicycle seat b) eraser}};

\node (p1) [box=green, minimum height=1.65cm, below of=q, yshift=-1.3cm, xshift=-3.2cm] {\textbf{P1}: Consider the following\\problem: \texttt{[Question]}\\List specific facts that\\seem most relevant to\\answering the question.};

\node (p2) [box=pastelblue, minimum height=1.65cm, below of=q, yshift=-1.3cm, xshift=1.9cm] {\textbf{P2}: Here are some facts that\\are relevant to the question\\I will ask you: \texttt{[Information]}\\Here is the question:\\\texttt{[Question]}};

\node (r) [box=purple, minimum height=1.0cm, below of=p2, yshift=-2.0cm] {\textbf{Information}: \textit{- LCD monitors are made of materials}\\
\textit{such as plastic and glass for the screen, alongside}\\
\textit{various internal electronic components.}\\
\textit{- Bicycle seats are typically made of materials}\\
\textit{like leather, vinyl, or foam padding.}\\
\textit{- Erasers are generally made of rubber or plastic.}};

\node (lm1) [box=pastelyellow, minimum width=1.9cm, minimum height=1.1cm, below of=p1, xshift=-0.2cm, yshift=-2.0cm] {\textbf{LM1}};

\node (lm2) [box=pastelyellow, minimum width=1.9cm, minimum height=1.1cm, right of=p2, xshift=3.5cm] {\textbf{LM2}};

\draw [dashed, thick, -{Latex[width=5pt]}] ([xshift=-2cm] q.south) -- ++(0,0) -- (p1.north);
\draw [dashed, thick, -{Latex[width=5pt]}] ([xshift=0.7cm] q.south) -- ++(0,0) -- (p2.north);
\draw [-{Triangle[width=7.5pt,length=5pt]}, line width=1.7pt] ([xshift=-0.2cm] p1.south) -- ++(0,0) -- (lm1.north);
\draw [dashed, thick, -{Latex[width=5pt]}] (r.north) -- ++(0,0) -- (p2.south);
\draw [-{Triangle[width=7.5pt,length=5pt]}, line width=1.7pt] (p2.east) -- ++(0,0) -- (lm2.west);
\draw [-{Triangle[width=7.5pt,length=5pt]}, line width=1.7pt] (lm1.east) -- ++(0,0) -- (r.west);
\end{tikzpicture}
    \caption{Overview of our PREP approach, a dual-instance prompting method using general and user-knowledge independent prompts (slightly simplified here). The first prompt (P1), combined with the question, directs LM1 to provide specific facts in its response that seem relevant to answering the question. The collected information is then combined with the original question and used to prompt LM2.}
    \label{fig:prompt-overview}
\end{figure*}

Our dual-instance prompting method, called PREP, is a simple technique that uses knowledge elicited from one instance of an LM to {\sc prep}are another instance of the LM for a QA task. It consists of two steps. The first step is knowledge elicitation, where we present the question to the first instance of the LM (LM1) and ask it to provide information relevant to answering the question. The second step is knowledge transfer, where the response from LM1 is copied and sent to the second instance of the LM (LM2) along with the question. LM2 now answers the question. Fig.~\ref{fig:prompt-overview} provides a visual representation of our PREP approach. 

This method is designed to be both general and user-knowledge independent; that is, it can be applied to a wide range of QA domains and tasks without requiring further prompt engineering, training, or other customizations that depend on the user's understanding of the task domain. In practice, our PREP prompting method offers a simple performance-enhancing method for querying LMs, without requiring extensive time investment in crafting specialized prompts. 
This makes it particularly attractive for scenarios where quick deployment and broad applicability are prioritized.

\section{Experimental Settings}

\subsection{Dataset Creation}
\label{sec:Dataset Creation}
We designed a nontrivial QA task, which asks LMs binary-choice questions about shared materials between objects. This task challenges LMs to consider detailed information about the possible materials that objects can be made of. To construct our dataset, we began with an extensive schematic dataset detailing the parts and material composition of primarily man-made objects \citep{an2024largelanguagemodelstool}. From this dataset, we identified triples of objects (\textit{$O_A$, $O_B$, $O_C$}) where objects \textit{$O_A$} and \textit{$O_B$} share a common material, while object \textit{$O_C$} does not share any material with \textit{$O_B$}. Examples are shown in Table \ref{Dataset examples}. We then manually selected 100 triples from this set and corrected any inaccuracies, creating curated test questions for our new dataset. The questions (\textbf{Q}) are formatted as follows:

\begin{quote}
Normally, which of the following is less likely to be at least partially made of a material that is a constituent of \textit{$O_B$}?

\;\; a) $O_A$ \hspace*{2mm} b) $O_C$
\end{quote}

\noindent
This phrasing is preferable to asking which artifact is \textit{more} likely to share materials with the target object, as it avoids the connotation that the amount of shared material is quantitatively significant for the correct answer.
To avoid any bias, we ensured there were an equal number of questions with `a)' and `b)' as the correct answers. 

The size of our shared material dataset is limited because of the difficulty of identifying triples of objects within our larger object database, where two objects share materials and one does not share any materials with one of those two. The variety of possible materials used in artifacts, their subtypes, and parts, often described in various ways, further complicates the process. Careful human curation is needed to ensure the quality of the dataset, and annotators need to agree on their understanding of the characterizations of the artifacts.

\begin{table*}[h!]
    \centering
    \begin{tabular}{cccc}
        \hline
        \boldmath{$O_A$} & \boldmath{$O_B$} & \boldmath{$O_C$} & \textbf{Shared between} \boldmath{$O_A$} \textbf{and} \boldmath{$O_B$}\\
        \hline
        windshield wiper & pop-up mosquito net & clear vase & metal\\
        bean bag chair & cape & spring doorstop & fabric\\
        trash can & dental braces & war flag & metal, plastic\\
        longbow & wrap (clothing) & razor blade & linen, polyester \\
        sky lantern & eco-friendly toothbrush & golf ball & bamboo \\
        \hline
    \end{tabular}
    \caption{Examples of triples (\textit{$O_A$, $O_B$, $O_C$}) from our curated dataset, illustrating shared material composition between $O_A$ and $O_B$, and the absence of shared materials between $O_B$ and $O_C$.}
    \label{Dataset examples}
\end{table*}

\subsection{Experiments}

We evaluate our PREP method by comparing it to various prompting techniques. The comparison includes methods with different levels of reliance on user knowledge, ranging from direct questioning, Plan and Solve (PS) prompting, and zero-shot Chain of Thought (CoT), to those incorporating user-knowledge dependent prompts tailored to the question. Specifically, in the user-knowledge dependent prompts, the LMs are asked to list the parts and materials of the objects mentioned in \textbf{Q} before providing an answer, rather than offering general information relevant to \textbf{Q}. We also examine other single-instance methods that utilize one or two prompts, both with and without these user-knowledge dependent prompts. To further investigate the impact of reinitializing the LM, we include variations where the information elicited from the LM is copied into the subsequent prompts within the same LM instance. Appendix~\ref{visual compar} presents examples of CoT, PS, and our proposed PREP method to illustrate their differences.
Additionally, we evaluate the performance of our PREP method on 500 questions from each of CSQA, StrategyQA, and OpenBookQA \citep{DBLP:journals/corr/abs-1809-02789}
. The specific prompts used in our tests are detailed in Appendix~\ref{specific prompts}.

Experiments are conducted using three language models: Phi-3 (14B) \citep{abdin2024phi3}, which is an efficient medium-sized model, Aya 23 (35B) \citep{aryabumi2024aya23}, which is trained from multilingual data containing 23 languages, and Command-R v0.1 (35B) \citep{c4ai2025commandr}, which is optimized for long context tasks such as retrieval-augmented generation (RAG). Our focus on medium-sized LMs comes from initial tests with smaller LMs (7B to 8B) on our shared material dataset. To our surprise, none of the tested methods was able to improve the performance of these models to a level differentiable from randomly guessing. Even when we provided substantial assistance, such as directly supplying the relevant material information to these LMs, their accuracy in answering \textbf{Q} barely exceeded random chance; so we believe it is very difficult to find a simple prompting method that works effectively on smaller LMs. Larger LMs (70B+) are not in the scope of this study because of our hardware constraints.

Based on these findings, our experiments focus on evaluating the performance of medium-sized LMs in controlled conditions. To reduce variability and highlight on the models' core capabilities, we set the temperature to 0, locking the output to be deterministic. The code used to conduct the experiments is publicly available on GitHub\footnote{\url{https://github.com/jasonyu48/PREP}}, and the dataset can be accessed via Hugging Face\footnote{\href{https://huggingface.co/datasets/jjjyu48/SharedMaterialQA}{https://huggingface.co/datasets/jjjyu48/SharedMaterial\linebreak QA}}.




\section{Results and Discussion}

\begin{table*}
\centering
\begin{tabular}{ccccccc}
\toprule
Method & \# Inst. & \# Messages & Copy & Command-R & Avg. Acc. \\
\midrule
ZS CoT & single & 1 & - & 59 & 65.33 \\
Direct & single & 1 & - & \textcolor{red}{54} &\textcolor{red}{58.67} \\
PS & single & 1 & - & \textcolor{blue}{\textbf{66}} & \textcolor{blue}{67.00} \\
\midrule
User Knowl-& single & 1 & - & \textcolor{red}{56} &\textcolor{blue}{65.67} \\
edge Indepen-& single & 2 & no & \textcolor{red}{46} & \textcolor{red}{59.67} \\
dent Elicita-& single & 2 & yes & \textcolor{red}{58} & \textcolor{red}{64.00} \\
tion & dual (ours) & 2 & yes & \textcolor{blue}{\textbf{66}} & \textcolor{blue}{\textbf{67.67}} \\
\midrule
User Knowl-& single & 1 & -  & \textcolor{blue}{64} & \textcolor{red}{62.33} \\
edge Depen-& single & 2 & no & \textcolor{blue}{62} &\textcolor{red}{60.67} \\
dent Elicita-& single & 2 & yes & \textcolor{blue}{60} & \textcolor{red}{63.00} \\
tion & dual (ours) & 2 & yes  & \textcolor{blue}{\textbf{74}} &\textcolor{blue}{\textbf{69.00}} \\
\bottomrule
\end{tabular}
\caption{Accuracy of prompting methods on our curated dataset. The far-right column reports the average accuracy across three models, while the second-to-right column shows results for the best-performing model. Zero-shot CoT (ZS CoT; first row in the table) serves as the baseline. Accuracy values higher than the baseline is highlighted in blue, while those lower are marked in red. For methods involving two user messages sent to the LM, a `yes' in the `Copy' column indicates that the second message includes a copy of the LM's response.}
\label{tab:accuracy-curated-dataset}
\end{table*}

\begin{table*}[h!]
\centering
\begin{tabular}{cccccccc}
\toprule
Method & \# Inst. & \# Messages & Copy & CSQA & StrategyQA & OpenBookQA \\
\midrule
ZS CoT & single & 1 & - & \textcolor{black}{73.20} & \textcolor{black}{72.07} &  \textcolor{black}{84.53} \\
Direct & single & 1 & - & \textcolor{blue}{76.93} & \textcolor{blue}{72.73} & \textcolor{blue}{86.67} \\
PS & single & 1 & - & \textcolor{red}{71.07} & \textcolor{blue}{74.80} &  \textcolor{blue}{84.80} \\
\midrule
User Knowl- & single & 1 & - & \textcolor{blue}{73.40} & \textcolor{blue}{74.73} &  \textcolor{blue}{86.00} \\
edge Indepen-& single & 2 & no & \textcolor{blue}{74.53} & \textcolor{blue}{74.53} &  \textcolor{red}{83.07} \\
dent Elicita-& single & 2 & yes & \textcolor{blue}{75.20} & \textcolor{blue}{74.47} &  \textcolor{blue}{85.27} \\
tion& dual (ours) & 2 & yes & \textcolor{blue}{\textbf{77.67}} & \textcolor{blue}{\textbf{76.44}} & \textcolor{blue}{\textbf{88.33}} \\
\bottomrule
\end{tabular}
\caption{Average accuracy of prompting methods on CommonsenseQA, StrategyQA, and OpenBookQA.}
\label{tab:accuracy-three-datasets}
\end{table*}


\noindent On all the tested datasets, the average accuracy of our dual-instance methods (averaged across the three models) consistently matches or exceeds the accuracy of all other tested methods within their user-knowledge dependency class. Table \ref{tab:accuracy-curated-dataset} highlights the accuracy of prompting methods on our curated dataset, reporting both the average accuracy across the three models and the results for the best-performing model. Notably, our method outperforms zero-shot CoT by approximately 4\% on the CSQA, StrategyQA, and OpenBookQA datasets, as shown in Table \ref{tab:accuracy-three-datasets}. Further breakdowns and results for individual models are provided in Appendix~\ref{detailed results}.

When comparing our method with the \textit{single inst. info. copied} method, we observe that repeating information in the context negatively impacts accuracy. This demonstrates the necessity of creating an additional instance of the LM.
Furthermore, when comparing our dual-instance method to those that do not copy information (especially \textit{single inst. 1 message} and \textit{single inst. 2 messages}), we find that although LMs receive similar contextual information in all these methods, the accuracy improves when the information is provided directly by the user. This shows that user-provided information can lead to a higher gain in performance over simply separating the elicitation of information with reasoning.
When testing Phi3 on StrategyQA, we observe that under CoT prompting, the model is unable to provide a definite answer to 21 questions. However, with our PREP method, this number decreases to 8, showing improved confidence in the model.
A sample conversation illustrating this improvement can be found in Appendix~\ref{becomes conf}. Overall, approximately 2\% of the tested questions remain unanswered by the LMs.

\section{Conclusions}

In this study, we introduced a dual-instance prompting method, PREP, aimed at improving the performance of LMs on QA tasks that require accessing relevant knowledge rather than following step-by-step reasoning. By employing two instances of LMs, one for information elicitation and the other for answering the question based on the elicited information, we demonstrate a notable improvement in accuracy compared to direct questioning and zero-shot CoT approaches across different models. It shows the efficacy of eliciting knowledge relevant to a question before asking for an answer. The method is particularly useful when using medium-sized LMs to answer questions dependent on commonsense knowledge. Our method can be applied without the need for domain-specific prompt engineering. Experimental results show that our dual-instance methods outperform single-instance methods. This underscores the value of (ostensibly) user-provided context in enhancing LM performance.

\section*{Limitations}

Our method is specifically designed for and has been tested on instruction fine-tuned language models. We are uncertain about its effectiveness on models that have not undergone instruction fine-tuning. Additionally, our evaluation primarily focused on question-answering tasks involving commonsense knowledge. Future work could explore the performance of our method in other domains.

The size of our shared material dataset is limited, as discussed in Section~\ref{sec:Dataset Creation} along with the associated challenges.



\bibliography{main}

\appendix
\section*{Appendix}
\label{sec:appendix}

\section{Model Details}
The language models are deployed offline using Ollama, available at \url{https://ollama.com/library}. Experiments are conducted on a single RTX4090.

\section{Prompt Details}
\label{specific prompts}
The exact prompts used in the experiment on our dataset are listed here. To adapt a user-knowledge independent prompt to the other datasets, we simply replace the term ``binary-choice'' with ``multiple-choice'' and change the last sentence of the prompt accordingly.

\begin{promptbox}{{Dual-Instance, User-Knowledge Independent, Information Copied (Ours)}}
\texttt{User's Message to LM1:}
\begin{quote}
Consider the following binary-choice problem:

\texttt{(}\textit{Insert the question \textbf{Q}}\texttt{)}

Please list specific facts that seem most relevant to answering the question. Do not answer the question, and do not include anything other than the list in your response.
\end{quote}
\texttt{User's Message to LM2:}
\begin{quote}
Here are some facts that are relevant to the question I will ask you:

\texttt{(}\textit{Insert the response of LM1}\texttt{)}

Here is the question:

\texttt{(}\textit{Insert the question \textbf{Q}}\texttt{)}

Consider the question based on common sense and the information. Clearly indicate the answer by saying `my answer is a)' or `my answer is b)' at the end of your response.
\end{quote}
\end{promptbox}

\begin{promptbox}{Direct Asking}
\texttt{User's Message:}
\begin{quote}
\texttt{(}\textit{Insert the question \textbf{Q}}\texttt{)}

Clearly indicate the answer by saying `my answer is a)' or `my answer is b)' at the end of your response.
\end{quote}
\end{promptbox}

\begin{promptbox}{Zero-shot CoT}
\texttt{User's Message:}
\begin{quote}
\texttt{(}\textit{Insert the question \textbf{Q}}\texttt{)}

Clearly indicate the answer by saying `my answer is a)' or `my answer is b)' at the end of your response.

\end{quote}
\texttt{Assistant's Message:}
\begin{quote}
Let's think step by step. \texttt{(}\textit{LM continues generation from here}\texttt{)}
\end{quote}
\end{promptbox}

\begin{promptbox}{Plan and Solve (PS) Prompting}
\texttt{User's Message:}
\begin{quote}
\texttt{(}\textit{Insert the question \textbf{Q}}\texttt{)}

Clearly indicate the answer by saying `my answer is a)' or `my answer is b)' at the end of your response.

\end{quote}
\texttt{Assistant's Message:}
\begin{quote}
Let's first understand the problem and devise a plan to solve the problem. Then, let's carry out the plan and solve the problem step by step. \texttt{(}\textit{LM continues generation from here}\texttt{)}
\end{quote}
\end{promptbox}

\begin{promptbox}{{Single-Instance, User-Knowledge Independent, One-Prompt}}
\texttt{User's Message:}
\begin{quote}
\texttt{(}\textit{Insert the question \textbf{Q}}\texttt{)}

Before giving your answer, please first list specific facts that seem most relevant to answering the question.

Clearly indicate the answer by saying `my answer is a)' or `my answer is b)' at the end of your response.
\end{quote}
\end{promptbox}

\begin{promptbox}{{Single-Instance, User-Knowledge Independent, Two-Prompts}}
\texttt{User's First Message:}
\begin{quote}
Consider the following binary-choice problem:

\texttt{(}\textit{Insert the question \textbf{Q}}\texttt{)}

Please list specific facts that seem most relevant to answering the question. Do not answer the question, and do not include anything other than the list in your response.
\end{quote}
\texttt{User's Second Message:}
\begin{quote}
Consider the question based on common sense and the information. Clearly indicate the answer by saying `my answer is a)' or `my answer is b)' at the end of your response.
\end{quote}
\end{promptbox}

\begin{promptbox}{{Single-Instance, User-Knowledge Independent,  Two-Prompts, Information Copied}}
\texttt{User's First Message:}
\begin{quote}
Consider the following binary-choice problem:

\texttt{(}\textit{Insert the question \textbf{Q}}\texttt{)}

Please list specific facts that seem most relevant to answering the question. Do not answer the question, and do not include anything other than the list in your response.
\end{quote}
\texttt{User's Second Message:}
\begin{quote}
Here are some facts that are relevant to the question:

\texttt{(}\textit{Insert the response from First Prompt}\texttt{)}

Consider the question based on common sense and the information. Clearly indicate the answer by saying `my answer is a)' or `my answer is b)' at the end of your response.
\end{quote}
\end{promptbox}

\begin{promptbox}{{Single-Instance, One-Prompt, User-Knowledge Dependent}}
\texttt{User's Message:}
\begin{quote}
\texttt{(}\textit{Insert the question \textbf{Q}}\texttt{)}

Before giving your answer, please first list the parts of $O_A$, $O_B$, and $O_C$, as well as the material of each part.

Clearly indicate the answer by saying `my answer is a)' or `my answer is b)' at the end of your response.
\end{quote}
\end{promptbox}

\begin{promptbox}{{Single-Instance, User-Knowledge Dependent,  Two-Prompts}}
\texttt{User's First Message:}
\begin{quote}
List the parts of $O_A$, $O_B$, and $O_C$, as well as the material of each part.
\end{quote}
\texttt{User's Second Message:}
\begin{quote}
\texttt{(}\textit{Insert the question \textbf{Q}}\texttt{)}

Consider the question based on common sense and the information. Clearly indicate the answer by saying `my answer is a)' or `my answer is b)' at the end of your response.
\end{quote}
\end{promptbox}

\begin{promptbox}{{Single-Instance, User-Knowledge Dependent, Two-Prompts, Information Copied}}
\texttt{User's First Message:}
\begin{quote}
List the parts of $O_A$, $O_B$, and $O_C$, as well as the material of each part.
\end{quote}
\texttt{User's Second Message:}
\begin{quote}
Here are some facts that are relevant to the question I will ask you:

\texttt{(}\textit{Insert the response from First Prompt}\texttt{)}

Here is the question:

\texttt{(}\textit{Insert the question \textbf{Q}}\texttt{)}

Consider the question based on common sense and the information. Clearly indicate the answer by saying `my answer is a)' or `my answer is b)' at the end of your response.
\end{quote}
\end{promptbox}

\begin{promptbox}{{Dual-Instance, User-Knowledge Dependent, Information Copied}}
\texttt{User's Message to LM1:}
\begin{quote}
List the parts of $O_A$, $O_B$, and $O_C$, as well as the material of each part.
\end{quote}
\texttt{User's Message to LM2:}
\begin{quote}
Here are some facts that are relevant to the question I will ask you:

\texttt{(}\textit{Insert the response of LM1}\texttt{)}

Here is the question:

\texttt{(}\textit{Insert the question \textbf{Q}}\texttt{)}

Consider the question based on common sense and the information. Clearly indicate the answer by saying `my answer is a)' or `my answer is b)' at the end of your response.
\end{quote}
\end{promptbox}


\section{Sample Conversations}
The following conversations are generated Phi3.
\subsection{Our PREP Method}
\noindent \textcolor{red}{\texttt{User's Message to LM1:}}

\vspace*{0.5em} \noindent Consider the following binary-choice problem:

\begin{tcolorbox}[width=0.5\textwidth, colback=white!90!black, frame empty]
\noindent Normally, which of the following is less likely to be at least partially made of a material that is a constituent of a magnifying glass?\\
a) doorstop b) contact lens
\end{tcolorbox}
 
\noindent Please list specific facts that seem most relevant to answering the question. Do not answer the question, and do not include anything other than the list in your response.
\vspace*{0.7em}\newline
\noindent \textcolor{blue}{\texttt{Response from LM1:}}

\begin{tcolorbox}[width=0.5\textwidth, colback=white!100!black]
\vspace*{0.5em} \noindent 1. Magnifying glasses are typically made from materials like plastic or glass with a convex shape for focusing light rays.\\
2. A doorstop is usually a solid object used to hold doors open or closed, often made of rubber, metal, wood, or stone.\\
3. Contact lenses are thin, curved pieces of clear material (usually plastic) designed to fit over the cornea for vision correction purposes.\\
4. The primary function of a magnifying glass is to focus light and enlarge images, while doorstops serve as physical barriers or holders for doors.\\
5. Contact lenses are made from materials like hydrogel or silicone that allow oxygen permeability and comfort on the eye surface.
\end{tcolorbox}

\noindent \textcolor{red}{\texttt{User's Message to LM2:}}

\vspace*{0.5em} \noindent Here are some facts that are relevant to the question I will ask you:
\begin{tcolorbox}[width=0.5\textwidth, colback=white!100!black]
\noindent 1. Magnifying glasses are typically made from materials like plastic or glass with a convex shape for focusing light rays.\\
2. A doorstop is usually a solid object used to hold doors open or closed, often made of rubber, metal, wood, or stone.\\
3. Contact lenses are thin, curved pieces of clear material (usually plastic) designed to fit over the cornea for vision correction purposes.\\
4. The primary function of a magnifying glass is to focus light and enlarge images, while doorstops serve as physical barriers or holders for doors.\\
5. Contact lenses are made from materials like hydrogel or silicone that allow oxygen permeability and comfort on the eye surface.
\end{tcolorbox}

\noindent Here is the question:

\begin{tcolorbox}[width=0.5\textwidth, colback=white!90!black, frame empty]
\noindent Normally, which of the following is less likely to be at least partially made of a material that is a constituent of a magnifying glass?\\
a) doorstop b) contact lens
\end{tcolorbox}

\noindent Consider the question based on common sense and the information. Clearly indicate the answer by saying `my answer is a)' or `my answer is b)' at the end of your response:
\vspace*{0.7em}\newline
\noindent \textcolor{blue}{\texttt{Response from LM2:}}

\vspace*{0.5em} \noindent Based on the given facts, we can deduce that magnifying glasses are typically made from materials like plastic or glass with a convex shape for focusing light rays. Contact lenses, however, are usually made from hydrogel or silicone to allow oxygen permeability and comfort on the eye surface. Therefore, it is less likely that contact lenses would be at least partially made of a material that is a constituent of a magnifying glass. So my answer is b).

\noindent \darkgreen{(Correct)}

\subsection{Direct}
\noindent \textcolor{red}{\texttt{User's Message:}}

\begin{tcolorbox}[width=0.5\textwidth, colback=white!90!black, frame empty]
\noindent Normally, which of the following is less likely to be at least partially made of a material that is a constituent of a magnifying glass?\\
a) doorstop b) contact lens
\end{tcolorbox}
 
\noindent Clearly indicate the answer by saying `my answer is a)' or `my answer is b)' at the end of your response.

\noindent \textcolor{blue}{\texttt{Assistant's Message:}}

\noindent My answer is a). A doorstop is typically made from materials like rubber, wood, or metal which are not constituents of a magnifying glass. On the other hand, contact lenses can be made from plastic polymers that have similar optical properties to those used in some types of magnifying glasses.

\noindent \noindent \darkred{(Wrong)}

\subsection{ZS CoT}
\noindent \textcolor{red}{\texttt{User's Message:}}

\begin{tcolorbox}[width=0.5\textwidth, colback=white!90!black, frame empty]
\noindent Normally, which of the following is less likely to be at least partially made of a material that is a constituent of a magnifying glass?\\
a) doorstop b) contact lens
\end{tcolorbox}
 
\noindent Clearly indicate the answer by saying `my answer is a)' or `my answer is b)' at the end of your response.

\noindent \textcolor{blue}{\texttt{Assistant's Message:}}

\noindent Let's think step by step. A magnifying glass typically consists of two main parts: the frame and the lens itself, which is usually made from a type of optical glass that has been shaped to focus light in a specific way. The material used for this purpose needs to have certain properties such as clarity, durability, and the ability to refract light effectively.

a) A doorstop is typically a small object designed to hold a door open or closed. It can be made from various materials like rubber, metal, wood, or plastic. These materials are chosen for their strength and stability rather than their optical properties. While it's possible that some part of the doorstop could include glass (for example, in decorative elements), this is not a primary function of the object.

b) A contact lens, on the other hand, must be made from materials that are safe for use with human eyes and have specific optical properties to correct vision. Contact lenses can indeed be made from types of glass or plastic (specifically designed polymers), which are also used in magnifying glasses due to their clarity and refractive abilities.

Therefore, my answer is a) doorstop.

\noindent \noindent \darkred{(Wrong)}

\subsection{PS Prompting}
\noindent \textcolor{red}{\texttt{User's Message:}}

\begin{tcolorbox}[width=0.5\textwidth, colback=white!90!black, frame empty]
\noindent Normally, which of the following is less likely to be at least partially made of a material that is a constituent of a magnifying glass?\\
a) doorstop b) contact lens
\end{tcolorbox}
 
\noindent Clearly indicate the answer by saying `my answer is a)' or `my answer is b)' at the end of your response.

\noindent \textcolor{blue}{\texttt{Assistant's Message:}}

\noindent Let's first understand the problem and devise a plan to solve the problem. Then, let's carry out the plan and solve the problem step by step.

The question asks us to compare two objects - doorstop and contact lens - in terms of their likelihood of being made from a material that is also used in making magnifying glasses. A magnifying glass typically uses a convex lens, which is usually made out of materials like plastic or glass with specific optical properties.

a) Doorstops can be made from various materials such as rubber, metal, wood, etc., but they are not commonly made from the same material used in making magnifying glasses (i.e., a type of lens).

b) Contact lenses, on the other hand, are often made out of plastic or silicone which can have similar optical properties to those found in some types of lenses like those used for magnifying glasses.

So, based on this analysis, doorstops (option a) are less likely to be at least partially made from the same material as a magnifying glass. 

Therefore, my answer is a).

\noindent \darkred{(Wrong)}

\subsection{Single-Instance, One-Prompt}
\noindent \textcolor{red}{\texttt{User's Message:}}

\begin{tcolorbox}[width=0.5\textwidth, colback=white!90!black, frame empty]
\noindent Normally, which of the following is less likely to be at least partially made of a material that is a constituent of a magnifying glass?\\
a) doorstop b) contact lens
\end{tcolorbox}
\noindent Before giving your answer, please first list specific facts that seem most relevant to answering the question.\\
\noindent Clearly indicate the answer by saying `my answer is a)' or `my answer is b)' at the end of your response.

\noindent \textcolor{blue}{\texttt{Assistant's Message:}}

\noindent To determine which item is less likely to be made from a material found in a magnifying glass, let's consider some facts:

1) A magnifying glass typically consists of two main parts - the lens and the frame. The lens itself is usually made of a type of transparent plastic or glass that has been shaped into a convex form to focus light rays for magnification purposes. This material must be optically clear, allowing light to pass through it without significant distortion.

2) A doorstop can come in various forms and materials such as rubber, metal, wood, or even stone. Its primary function is not related to the manipulation of light but rather to prevent a door from closing.

3) Contact lenses are small, thin pieces that fit directly onto the eye's surface. They need to be transparent for vision correction and often made from hydrogel or silicone materials which allow oxygen permeability and comfort when worn on the eye. However, they also require a certain level of optical clarity similar to magnifying glasses as they must let light pass through them without distortion.

4) The primary function of both contact lenses and doorstops is not related to focusing or manipulating light but rather correcting vision in the case of contacts, and preventing a door from closing for doorstops.

5) Magnifying glasses are specifically designed with optical properties that allow them to magnify objects when held at certain distances. This requires materials like plastic or glass which can be shaped into lenses.

Based on these facts, it seems less likely that a doorstop would contain material similar to what is found in a magnifying glass because its function and design do not require the optical properties of such materials. 

My answer is a).

\noindent \darkred{(Wrong)}

\subsection{Single-Instance, Two-Prompts}
\noindent \textcolor{red}{\texttt{User's Message:}}
\newline
\noindent Consider the following binary-choice problem:
\begin{tcolorbox}[width=0.5\textwidth, colback=white!90!black, frame empty]
\noindent Normally, which of the following is less likely to be at least partially made of a material that is a constituent of a magnifying glass?\\
a) doorstop b) contact lens
\end{tcolorbox}
\noindent Please list specific facts that seem most relevant to answering the question. Do not answer the question, and do not include anything other than the list in your response.

\noindent \textcolor{blue}{\texttt{Assistant's Message:}}

1. Magnifying glasses are typically made from materials like plastic or glass with a convex shape for focusing light rays.

2. A doorstop is usually a solid object used to hold doors open or closed, often made of rubber, metal, wood, or stone.

3. Contact lenses are thin, curved pieces of clear material (usually plastic) designed to fit over the cornea for vision correction purposes.

4. The primary function of a magnifying glass is to focus light and enlarge images, while doorstops serve as physical barriers or holders for doors.

5. Contact lenses are made from materials like hydrogel or silicone that allow oxygen permeability and comfort on the eye surface.

\noindent \textcolor{red}{\texttt{User's Message:}}
\newline
\noindent Consider the question based on common sense and the information. Clearly indicate the answer by saying `my answer is a)' or `my answer is b)' at the end of your response.

\noindent \textcolor{blue}{\texttt{Assistant's Message:}}

1. A magnifying glass typically uses materials like plastic or glass with specific optical properties to focus light rays, which are not commonly used in doorstops.

2. Doorstops serve as physical barriers and do not require specialized optical properties for their functioning.

3. Contact lenses need to be transparent and have a certain level of oxygen permeability, but they don't necessarily focus light like magnifying glasses.

4. The materials used in contact lenses are chosen based on biocompatibility with the eye rather than optical properties for image enlargement or focusing.

5. Doorstops do not require any specific material that can manipulate light to create a magnified view of objects, unlike magnifying glasses which need such properties.

6. The primary function of doorstops is unrelated to the manipulation of light and images.

7. Contact lenses are designed for direct contact with human skin (eye), while magnifying glasses do not come into contact with skin but rather interact with light passing through them.

8. Doorstops can be made from a variety of materials, including those that don't have any optical properties at all.

9. The primary function of doorstops is to hold doors open or closed, which does not involve the manipulation of light and images like magnifying glasses do.

10. Magnifying glasses are specifically designed for their ability to focus light rays due to their shape and material properties, while contact lenses have a different purpose altogether.\\
my answer is b)

\noindent \darkgreen{(Correct)}

\subsection{Single-Instance,
Two-Prompts, Information Copied}
\noindent \textcolor{red}{\texttt{User's Message:}}
\newline
\noindent Consider the following binary-choice problem:
\begin{tcolorbox}[width=0.5\textwidth, colback=white!90!black, frame empty]
\noindent Normally, which of the following is less likely to be at least partially made of a material that is a constituent of a magnifying glass?\\
a) doorstop b) contact lens
\end{tcolorbox}
\noindent Please list specific facts that seem most relevant to answering the question. Do not answer the question, and do not include anything other than the list in your response.

\noindent \textcolor{blue}{\texttt{Assistant's Message:}}

\begin{tcolorbox}[width=0.5\textwidth, colback=white!100!black]
\vspace*{0.5em} \noindent 1. Magnifying glasses are typically made from materials like plastic or glass with a convex shape for focusing light rays.\\
2. A doorstop is usually a solid object used to hold doors open or closed, often made of rubber, metal, wood, or stone.\\
3. Contact lenses are thin, curved pieces of clear material (usually plastic) designed to fit over the cornea for vision correction purposes.\\
4. The primary function of a magnifying glass is to focus light and enlarge images, while doorstops serve as physical barriers or holders for doors.\\
5. Contact lenses are made from materials like hydrogel or silicone that allow oxygen permeability and comfort on the eye surface.
\end{tcolorbox}

\noindent \textcolor{red}{\texttt{User's Message:}}
\newline
\noindent Here are some facts that are relevant to the question:
\begin{tcolorbox}[width=0.5\textwidth, colback=white!100!black]
\vspace*{0.5em} \noindent 1. Magnifying glasses are typically made from materials like plastic or glass with a convex shape for focusing light rays.\\
2. A doorstop is usually a solid object used to hold doors open or closed, often made of rubber, metal, wood, or stone.\\
3. Contact lenses are thin, curved pieces of clear material (usually plastic) designed to fit over the cornea for vision correction purposes.\\
4. The primary function of a magnifying glass is to focus light and enlarge images, while doorstops serve as physical barriers or holders for doors.\\
5. Contact lenses are made from materials like hydrogel or silicone that allow oxygen permeability and comfort on the eye surface.
\end{tcolorbox}

Consider the question based on common sense and the information. Clearly indicate the answer by saying `my answer is a)' or `my answer is b)' at the end of your response.

\noindent \textcolor{blue}{\texttt{Assistant's Message:}}

\noindent Based on the facts provided, it's clear that doorstops are typically made from materials like rubber, metal, wood, or stone and do not have any relation to magnifying glasses which are usually made from plastic or glass with a convex shape for focusing light rays. On the other hand, contact lenses could potentially be made of a material that is also used in some types of magnifying glasses (plastic). However, it's important to note that not all materials used in magnifying glasses are suitable for use in contact lenses due to safety and comfort considerations. Therefore, considering the likelihood based on common usage and manufacturing practices:\\
my answer is a) doorstop

\noindent \darkred{(Wrong)}

\section{Examples Illustrating Confidence Improvements with PREP}
\label{becomes conf}
\subsection{Example of a Non-Definite Answer Under ZS CoT Prompting}
\noindent \textcolor{red}{\texttt{User's Message:}}

\begin{tcolorbox}[width=0.5\textwidth, colback=white!90!black, frame empty]
\noindent Is cactus fruit an important menu item for a restaurant inspired by Cuauhtémoc?
\end{tcolorbox}
 
\noindent Clearly indicate the answer by saying `my answer is yes' or `my answer is no' at the end of your response.

\noindent \textcolor{blue}{\texttt{Assistant's Message:}}

\noindent Let's think step by step. Cactus fruit, also known as prickly pear, has been a part of Mexican cuise for centuries and it could be considered an important menu item in a restaurant inspired by Cuauhtémoc due to its historical significance and cultural relevance. However, the importance of this ingredient would depend on various factors such as the target audience's preferences, availability, and how well it can be incorporated into dishes that align with the overall theme of the restaurant.

Cuauhtémoc was an Aztec emperor who ruled during a time when cactus fruit might have been more commonly consumed than today. If the restaurant is aiming to provide authentic pre-Hispanic Mexican cuisine, then including cactus fruit could be seen as important for historical accuracy and cultural representation.

On the other hand, if the restaurant's focus is on modern interpretations of traditional dishes or a fusion menu that caters to contemporary tastes, it might not prioritize this ingredient unless there are creative ways to incorporate it into their offerings. Additionally, cactus fruit may not be as readily available in all regions and could pose challenges for sourcing and consistency of supply.

Considering these factors, my answer is that the importance of cactus fruit on a restaurant menu inspired by Cuauhtémoc would depend on the specific vision and concept of the establishment. If it aligns with their theme and they can source it reliably, then yes, it could be an important item; otherwise, no, it might not be as crucial to include.

My answer is: It depends.

\noindent \darkred{(Ambiguous)}

\subsection{Example of a Definite Answer Achieved with PREP}
\noindent \textcolor{red}{\texttt{User's Message to LM1:}}

\vspace*{0.5em} \noindent Consider the following question:

\begin{tcolorbox}[width=0.5\textwidth, colback=white!90!black, frame empty]
\noindent Is cactus fruit an important menu item for a restaurant inspired by Cuauhtémoc?
\end{tcolorbox}
 
\noindent Please list specific facts that seem most relevant to answering the question. Do not answer the question, and do not include anything other than the list in your response.
\vspace*{0.7em}\newline
\noindent \textcolor{blue}{\texttt{Response from LM1:}}

\begin{tcolorbox}[width=0.5\textwidth, colback=white!100!black]
\vspace*{0.5em} \noindent 1. Cuisine of Mexico: Understanding the traditional Mexican cuis vice is essential as it will help determine if cactus fruit fits into the menu items typically served in a restaurant inspired by Cuauhtémoc, which is located in Mexico City.\\
2. Availability and popularity of nopal (cactus) dishes: Nopales are commonly used in Mexican cuisine; they can be found in salads, tacos, quesadillas, etc., so it's likely that cactus fruit could also be a part of the menu.\\
3. Culinary use of nopal fruit (tunas): The fruits are edible and used to make jams, jellies, juices, or eaten fresh in Mexico; this indicates they might be included on the menu.\\
4. Local produce: If cactus is grown locally around Cuauhtémoc, it would likely be a part of local cuisine and thus could feature prominently on the restaurant's menu.\\
5. Restaurant theme or concept: A restaurant inspired by Cuauhtémoc might focus on traditional Mexican ingredients; cactus fruit is native to Mexico and may fit into this theme.\\
6. Customer preferences in the area: If customers in Cuauhtémoc are open to trying new, local produce like nopal fruit, it could be a popular menu item.\\
7. Dietary trends or health benefits of cactus fruit: Cactus fruits are rich in antiocks and vitamins; if the restaurant is targeting health-conscious customers, this might make them an important menu item.
\end{tcolorbox}

\noindent \textcolor{red}{\texttt{User's Message to LM2:}}

\vspace*{0.5em} \noindent Here are some facts that are relevant to the question I will ask you:
\begin{tcolorbox}[width=0.5\textwidth, colback=white!100!black]
\noindent 1. Cuisine of Mexico: Understanding the traditional Mexican cuis vice is essential as it will help determine if cactus fruit fits into the menu items typically served in a restaurant inspired by Cuauhtémoc, which is located in Mexico City.\\
2. Availability and popularity of nopal (cactus) dishes: Nopales are commonly used in Mexican cuisine; they can be found in salads, tacos, quesadillas, etc., so it's likely that cactus fruit could also be a part of the menu.\\
3. Culinary use of nopal fruit (tunas): The fruits are edible and used to make jams, jellies, juices, or eaten fresh in Mexico; this indicates they might be included on the menu.\\
4. Local produce: If cactus is grown locally around Cuauhtémoc, it would likely be a part of local cuisine and thus could feature prominently on the restaurant's menu.\\
5. Restaurant theme or concept: A restaurant inspired by Cuauhtémoc might focus on traditional Mexican ingredients; cactus fruit is native to Mexico and may fit into this theme.\\
6. Customer preferences in the area: If customers in Cuauhtémoc are open to trying new, local produce like nopal fruit, it could be a popular menu item.\\
7. Dietary trends or health benefits of cactus fruit: Cactus fruits are rich in antiocks and vitamins; if the restaurant is targeting health-conscious customers, this might make them an important menu item.
\end{tcolorbox}

\noindent Here is the question:

\begin{tcolorbox}[width=0.5\textwidth, colback=white!90!black, frame empty]
\noindent Is cactus fruit an important menu item for a restaurant inspired by Cuauhtémoc?
\end{tcolorbox}

\noindent Consider the question based on common sense and the information. Clearly indicate the answer by saying `my answer is yes' or `my answer is no' at the end of your response.
\vspace*{0.7em}\newline
\noindent \textcolor{blue}{\texttt{Response from LM2:}}

\vspace*{0.5em} \noindent Based on the provided facts, it seems that cactus fruit could indeed be an important menu item for a restaurant inspired by Cuauhtémoc. The traditional Mexican cuisine often includes nopales (cactus pads), and their fruits are also edible and used in various dishes like jams, jellies, juices, or eaten fresh. If the restaurant is focusing on local ingredients native to Mexico, cactus fruit could fit into this theme. Additionally, if customers in Cuauhtémoc are open to trying new produce and the health benefits of cactus fruits might appeal to them, it further supports their inclusion on the menu. Therefore, my answer is yes.

\noindent \darkgreen{(Correct and definite)}

\onecolumn
\section{Detailed Experiment Data for Each Model}
\label{detailed results}

\begin{table}[h!]
\centering
\begin{tabularx}{\textwidth}{lc@{\hskip 0.08in}cccccc}
\toprule
Method & \# Inst. & \# Messages & Copy & Phi-3 & Aya & Command-R & Avg. Diff \\
\midrule
ZS CoT & single & 1 & - & \textbf{78±4.1\%} & 59±4.9\% & 59±4.9\% & \textcolor{black}{\phantom{-}0.00} \\
Direct Asking & single & 1 & - & 67±4.7\% & 55±5\% & 54±5\% & \textcolor{red}{-6.67} \\
PS prompting & single & 1 & - & 72±4.5\% & 63±4.8\% & 66±4.7\% & \textcolor{blue}{\phantom{-}1.67} \\
\midrule
User Knowl-& single & 1 & - & 73±4.4\% & 68±4.7\% & 56±5\% & \textcolor{blue}{\phantom{-}0.33} \\
edge Indepen-& single & 2 & no & 71±4.5\% & 62±4.9\% & 46±5\% & \textcolor{red}{-5.67} \\
dent Elicita-& single & 2 & yes & 70±4.6\% & 64±4.8\% & 58±4.9\% & \textcolor{red}{-1.33} \\
tion & dual (ours) & 2 & yes & 70±4.6\% & 67±4.7\% & 66±4.7\% & \textcolor{blue}{\phantom{-}2.33} \\
\midrule
User Knowl-& single & 1 & - & 65±4.8\% & 58±4.9\% & 64±4.8\% & \textcolor{red}{-3.00} \\
edge Depen- & single & 2 & no & 60±4.9\% & 60±4.9\% & 62±4.9\% & \textcolor{red}{-4.67} \\
dent Elicita-& single & 2 & yes & 61±4.9\% & 68±4.7\% & 60±4.9\% & \textcolor{red}{-2.33} \\
tion & dual (ours) & 2 & yes & 62±4.9\% & \textbf{71±4.5\%} & \textbf{74±4.4\%} & \textcolor{blue}{\phantom{-}3.67} \\
\bottomrule
\end{tabularx}
\caption{Accuracy of prompting methods tested on our curated dataset. Zero-shot CoT (ZS CoT; first line in the table) is set as the baseline. Since multiple models are tested, we calculate the difference between the average accuracy of each prompting method and the average accuracy of ZS CoT to help examine the results. If the average accuracy is higher than the average accuracy of CoT, the Avg. Diff is blue. Otherwise, it is red.}
\end{table}


\begin{table}[h!]
\centering
\begin{tabularx}{\textwidth}{l@{\hskip 0.1in}c@{\hskip 0.08in}c@{\hskip 0.08in}ccccc}
\toprule
Method & \# Inst. & \# Messages & Copy & Phi-3 & Aya & Command-R & Avg. Diff \\
\midrule
ZS CoT & single & 1 & - & \textbf{74.2±2\%} & 72.4±2\% & 73±2\% & \textcolor{black}{0.00} \\
Direct & single & 1 & - & 73.2±2\% & \textbf{83±1.7\%} & 74.6±1.9\% & \textcolor{blue}{3.73} \\
PS & single & 1 & - & 72.6±2\% & 71.4±2\% & 69.2±2.1\% & \textcolor{red}{-2.13} \\
\midrule
User Knowl- & single & 1 & - & 71.6±2\% & 78±1.9\% & 70.6±2\% & \textcolor{blue}{0.20} \\
edge Indepen-& single & 2 & no & 69.8±2.1\% & 81.6±1.7\% & 72.2±2\% & \textcolor{blue}{1.33} \\
dent Elicita-& single & 2 & yes & 71.4±2\% & 82.2±1.7\% & 72±2\% & \textcolor{blue}{2.00} \\
tion& dual (ours) & 2 & yes & 73.4±2\% & 82.6±1.7\% & \textbf{77±1.9\%} & \textcolor{blue}{4.47} \\
\bottomrule
\end{tabularx}
\caption{Accuracy of prompting methods tested on CommonsenseQA}
\end{table}


\begin{table}[h!]
\centering
\begin{tabularx}{\textwidth}{l@{\hskip 0.1in}c@{\hskip 0.08in}c@{\hskip 0.08in}ccccc}
\toprule
Method & \# Inst. & \# Messages & Copy & Phi-3 & Aya & Command-R & Avg. Diff \\
\midrule
ZS CoT & single & 1 & - & 59.2±2.2\% & 78.6±1.8\% & 78.4±1.8\% & \textcolor{black}{0.00} \\
Direct & single & 1 & - & 63±2.2\% & 76.4±1.9\% & 78.8±1.8\% & \textcolor{blue}{0.67} \\
PS & single & 1 & - & 67±2.1\% & 77.2±1.9\% & 80.2±1.8\% & \textcolor{blue}{2.73} \\
\midrule
User Knowl- & single & 1 & - & 64.8±2.1\% & \textbf{80.4±1.8\%} & 79±1.8\% & \textcolor{blue}{2.67} \\
edge Indepen-& single & 2 & no & 66.4±2.1\% & 79.4±1.8\% & 77.8±1.9\% & \textcolor{blue}{2.47} \\
dent Elicita-& single & 2 & yes & 65±2.1\% & 80±1.8\% & 78.4±1.8\% & \textcolor{blue}{2.40} \\
tion& dual (ours) & 2 & yes  & \textbf{70.8±2\%} & 77.8±1.9\% & \textbf{80.4±1.8\%} & \textcolor{blue}{4.27} \\
\bottomrule
\end{tabularx}
\caption{Accuracy of prompting methods tested on StrategyQA}
\end{table}

\newpage
\begin{table}[h!]
\centering
\begin{tabularx}{\textwidth}{l@{\hskip 0.09in}c@{\hskip 0.07in}c@{\hskip 0.08in}ccccc}
\toprule
Method & \# Inst. & \# Messages & Copy & Phi-3 & Aya & Command-R & Avg. Diff \\
\midrule
ZS CoT & single & 1 & - & 89.2±1.4\% & 79.4±1.8\% & 85±1.6\% & \textcolor{black}{0.00} \\
Direct & single & 1 & - & 90.4±1.3\% & 87.4±1.5\% & 82.2±1.7\% & \textcolor{blue}{2.13} \\
PS & single & 1 & - & \textbf{92.2±1.2\%} & 80.2±1.8\% & 82±1.7\% & \textcolor{blue}{0.27} \\
\midrule
User Knowl- & single & 1 & - & 90±1.3\% & 86.8±1.5\% & 81.2±1.7\% & \textcolor{blue}{1.47} \\
edge Indepen-& single & 2 & no & 81.6±1.7\% & 85±1.6\% & 82.6±1.7\% & \textcolor{red}{-1.47} \\
dent Elicita-& single & 2 & yes & 87.2±1.5\% & 85.6±1.6\% & 83±1.7\% & \textcolor{blue}{0.73} \\
tion& dual (ours) & 2 & yes  & 88.2±1.4\% & \textbf{90.2±1.3\%} & \textbf{86.6±1.5\%} & \textcolor{blue}{3.80} \\
\bottomrule
\end{tabularx}
\caption{Accuracy of prompting methods tested on OpenBookQA}
\end{table}

\noindent We believe average accuracy (averaged across models) can better reflect the performance of prompting methods, because it is less model-dependent. Admittedly, our dual instance methods do not always perform the best for every model on every dataset, but in most situations where it is not the best method, its performance is close to the best-performing method. However, other methods that perform the best for a certain model on a certain dataset do not have consistent performance across all models. Our method remains a good choice for users who cannot perform model-specific tests.

\section{Illustrative Comparison of Prompting Techniques}
\label{visual compar}
\begin{figure}[h!]
\centering
\begin{tikzpicture}
    \node[anchor=north west] (ZSQbox) at (-0.7, 0) {
        \begin{tcolorbox}[colback=white, colframe=black, width=0.47\textwidth, top=2pt, bottom=2pt, left=2pt, right=2pt]
        \textbf{Q:} Normally, which of the following is less likely to be at least partially made of a material that is a constituent of a magnifying glass?\\
        a) doorstop b) contact lens\\
        Clearly indicate the answer by saying `my answer is a)' or `my answer is b)' at the end of your response.\\
        \par\vspace{0.15em}
        \textbf{A:} \darkred{\textit{Let's think step by step.}}
        \end{tcolorbox}
    };
    \node[above=0.1mm of ZSQbox] (ZStitle) {\textbf{Zero-shot CoT}};
    \node[anchor=north west] (ZSAbox) at (-0.7, -4.19){
        \begin{tcolorbox}[colback=white, colframe=black, width=0.47\textwidth, top=2pt, bottom=3pt, left=2pt, right=2pt,
        enhanced, 
        overlay={
        \node at (frame.south west) [xshift=4.9cm, yshift=19pt] {\tikzxmark};
    }]
        A magnifying glass typically consists of two main parts: the frame and the lens itself, which is usually made from a type of optical glass that has been shaped to focus light in a specific way. The material used for this purpose needs to have certain properties such as clarity, durability, and the ability to refract light effectively.\\
        \par\vspace{0.1em}
        a) A doorstop is typically a small object designed to hold a door open or closed. It can be made from various materials like rubber, metal, wood, or plastic. These materials are chosen for their strength and stability rather than their optical properties. While it's possible that some part of the doorstop could include glass (for example, in decorative elements), this is not a primary function of the object.\\
        b) A contact lens, on the other hand, must be made from materials that are safe for use with human eyes and have specific optical properties to correct vision. Contact lenses can indeed be made from types of glass or plastic (specifically designed polymers), which are also used in magnifying glasses due to their clarity and refractive abilities.\\
        \par\vspace{0.1em}
        Therefore, my answer is a) doorstop.
    \end{tcolorbox}
    };
    
    \node[anchor=north west] (PSQbox) at (7.5, 0) {
        \begin{tcolorbox}[colback=white, colframe=black, width=0.47\textwidth, top=2pt, bottom=2pt, left=2pt, right=2pt]
        \textbf{Q:} Normally, which of the following is less likely to be at least partially made of a material that is a constituent of a magnifying glass?\\
        a) doorstop b) contact lens\\
        Clearly indicate the answer by saying `my answer is a)' or `my answer is b)' at the end of your response.\\
        \par\vspace{0.15em}
        \textbf{A:} \darkred{\textit{Let's first understand the problem and devise a plan to solve the problem. Then, let's carry out the plan and solve the problem step by step.}}
        \end{tcolorbox}
    };
    \node[above=0.1mm of PSQbox] (PStitle) {\textbf{Plan-and-Solve (PS)}};
    \node[anchor=north west] (PSAbox) at (7.5, -4.8){
        \begin{tcolorbox}[colback=white, colframe=black, width=0.47\textwidth, top=2pt, bottom=3pt, left=2pt, right=2pt,
       enhanced, 
        overlay={
        \node at (frame.south west) [xshift=3.75cm, yshift=18.5pt] {\tikzxmark};
    }]
        The question asks us to compare two objects - doorstop and contact lens - in terms of their likelihood of being made from a material that is also used in making magnifying glasses. A magnifying glass typically uses a convex lens, which is usually made out of materials like plastic or glass with specific optical properties.\\
        \par\vspace{0.08em}
        Now let's consider the two options:\\
        a) Doorstops can be made from various materials such as rubber, metal, wood, etc., but they are not commonly made from the same material used in making magnifying glasses (i.e., a type of lens).\\
        b) Contact lenses, on the other hand, are often made out of plastic or silicone which can have similar optical properties to those found in some types of lenses like those used for magnifying glasses.\\
        \par\vspace{0.08em}
        So, based on this analysis, doorstops (option a) are less likely to be at least partially made from the same material as a magnifying glass.\\
        Therefore, my answer is a).
    \end{tcolorbox}
    };

    \node[draw, fill=blue!10, rectangle, rounded corners, minimum width=1cm, minimum height=0.5cm, below=2.1mm of ZSQbox] (ZSLLM) {LM};
    \draw[->, line width=0.45mm] (3.195, -3.1) -- (ZSLLM.north);
    \draw[->, line width=0.45mm] (ZSLLM.south) -- (3.195, -4.31);
    \node[draw, fill=blue!10, rectangle, rounded corners, minimum width=1cm, minimum height=0.5cm, below=2.1mm of PSQbox] (PSLLM) {LM};
    \draw[->, line width=0.45mm] (11.4, -3.72) -- (PSLLM.north);
    \draw[->, line width=0.45mm] (PSLLM) -- (11.4, -4.93);

    \draw[dotted, gray, line width=0.4mm, dash pattern=on 1.3mm off 1.3mm]
            ([xshift=0.2cm, yshift=0.1cm]ZSQbox.north east) -- ([xshift=-0.2cm, yshift=-10cm]PSQbox.south west);

\end{tikzpicture}
\end{figure}

\begin{figure}[h!]
\centering
\begin{tikzpicture}
    \node[anchor=north west] (PREPQbox) at (-0.7, 0) {
        \begin{tcolorbox}[colback=white, colframe=black, width=0.47\textwidth, top=2pt, bottom=2pt, left=2pt, right=2pt]
        \darkred{Consider the following binary-choice problem:}\\
        \par\vspace{0.15em}
        \textbf{Q:} Normally, which of the following is less likely to be at least partially made of a material that is a constituent of a magnifying glass?\\
        a) doorstop b) contact lens\\
        Clearly indicate the answer by saying `my answer is a)' or `my answer is b)' at the end of your response.\\
        \par\vspace{0.15em}
        \darkred{\textit{Please list specific facts that seem most relevant to answering the question. Do not answer the question, and do not include anything other than the list in your response.}}
        \end{tcolorbox}
    };
    \node[above=0.1mm of PREPQbox] (PREPtitle) {\textbf{Our PREP Method}};
    \node[anchor=north west] (PREPAbox) at (-0.7, -5.5){
        \begin{tcolorbox}[colback=white, colframe=black, width=0.47\textwidth, top=2pt, bottom=3pt, left=2pt, right=2pt]
        1. Magnifying glasses are typically made from materials like plastic or glass with a convex shape for focusing light rays.\\
        2. A doorstop is usually a solid object used to hold doors open or closed, often made of rubber, metal, wood, or stone.\\
        3. Contact lenses are thin, curved pieces of clear material (usually plastic) designed to fit over the cornea for vision correction purposes.\\
        4. The primary function of a magnifying glass is to focus light and enlarge images, while doorstops serve as physical barriers or holders for doors.\\
        5. Contact lenses are made from materials like hydrogel or silicone that allow oxygen permeability and comfort on the eye surface.
        \end{tcolorbox}
    };
    \node[anchor=north west] (PREPQbox2) at (7.5, -5.5) {
        \begin{tcolorbox}[colback=white, colframe=black, width=0.475\textwidth, top=2pt, bottom=2pt, left=2pt, right=2pt]
        \darkred{Here are some facts that are relevant to the question I will ask you:}\\
        \par\vspace{0.15em}
        1. Magnifying glasses are typically made from materials like plastic or glass with a convex shape for focusing light rays.\\
        2. A doorstop is usually a solid object used to hold doors open or closed, often made of rubber, metal, wood, or stone.\\
        3. Contact lenses are thin, curved pieces of clear material (usually plastic) designed to fit over the cornea for vision correction purposes.\\
        4. The primary function of a magnifying glass is to focus light and enlarge images, while doorstops serve as physical barriers or holders for doors.\\
        5. Contact lenses are made from materials like hydrogel or silicone that allow oxygen permeability and comfort on the eye surface.\\
        \par\vspace{0.15em}
        \darkred{Here is the question:}\\
        \par\vspace{0.15em}
        \textbf{Q:} Normally, which of the following is less likely to be at least partially made of a material that is a constituent of a magnifying glass?\\
        a) doorstop b) contact lens\\
        \par\vspace{0.15em}
        \darkred{Consider the question based on common sense and the information.} Clearly indicate the answer by saying `my answer is a)' or `my answer is b)' at the end of your response:
        \end{tcolorbox}
    };
    \node[anchor=north west] (PREPAbox2) at (7.5, -1){
        \begin{tcolorbox}[colback=white, colframe=black, width=0.475\textwidth, top=2pt, bottom=3pt, left=2pt, right=2pt,
       enhanced, 
        overlay={
        \node at (frame.south east) [xshift=-5.5mm, yshift=17.5
        pt] {\tikzcmark};
    }]
        Based on the given facts, we can deduce that magnifying glasses are typically made from materials like plastic or glass with a convex shape for focusing light rays. Contact lenses, however, are usually made from hydrogel or silicone to allow oxygen permeability and comfort on the eye surface. Therefore, it is less likely that contact lenses would be at least partially made of a material that is a constituent of a magnifying glass. So my answer is b).
    \end{tcolorbox}
    };

    \node[draw, fill=blue!10, rectangle, rounded corners, minimum width=1cm, minimum height=0.5cm, below=2.1mm of PREPQbox] (PREPLLM) {LM1};
    \draw[->, line width=0.45mm] (3.195, -4.4) -- (PREPLLM.north);
    \draw[->, line width=0.45mm] (PREPLLM.south) -- (3.195, -5.63);
    \node[draw, fill=blue!10, rectangle, rounded corners, minimum width=1cm, minimum height=0.5cm, below=3.48mm of PREPAbox2] (PREPLLM2) {LM2};
    \draw[->, line width=0.45mm] (PREPLLM2.north) -- (11.44, -4.12);
    \draw[->, line width=0.45mm] (11.438, -5.63) -- (PREPLLM2.south);
    \draw[->, dashed, line width=0.45mm] (6.92, -7.8) -- (7.63, -7.8);
    
\end{tikzpicture}
\end{figure}

\end{document}